\pdfoutput=1

\documentclass[11pt]{article}

\usepackage[preprint]{acl}
\usepackage{tcolorbox}
\usepackage{graphicx}
\usepackage{color}

\usepackage{soul}
\usepackage{url}
\usepackage[utf8]{inputenc}
\usepackage{caption}
\usepackage{amsmath}
\usepackage{amsthm}
\usepackage{booktabs}
\usepackage{algorithm}
\usepackage{algorithmic}
\usepackage[switch]{lineno}
\usepackage{amssymb}
\usepackage{natbib}
\usepackage{arydshln}
\usepackage{multirow}
\definecolor{lightgray}{gray}{0.60}
\usepackage{tcolorbox} 
\newcommand{\hil}[1]{\textcolor{lightgray}{#1}}
\usepackage{times}
\usepackage{latexsym}

\usepackage[T1]{fontenc}

\usepackage[utf8]{inputenc}

\usepackage{microtype}

\usepackage{inconsolata}

\usepackage{graphicx}
\usepackage{algorithm}
\usepackage{algorithmic}

\usepackage{pifont}

\usepackage{multirow}
\usepackage{graphicx}
\usepackage{booktabs}
\usepackage{makecell}
\usepackage{xcolor}
\usepackage{vcell}
\usepackage{colortbl}

\usepackage{todonotes}

\title{\textsc{SynDec}: A Synthesize-then-Decode Approach for Arbitrary Textual Style Transfer via Large Language Models}

\author{
 \textbf{Han Sun\textsuperscript{1}}\thanks{Equal contribution.},
 \textbf{Zhen Sun\textsuperscript{2}}\footnotemark[1],
 \textbf{Zongmin Zhang\textsuperscript{2}}\footnotemark[1],
 \textbf{Linzhao Jia\textsuperscript{1}},
 \textbf{Wei Shao\textsuperscript{3}},
 \textbf{Min Zhang\textsuperscript{1}}\thanks{Corresponding author (\texttt{zhangmin@sei.ecnu.edu.cn})}
\\
\\
 \textsuperscript{1}East China Normal University \\
 \textsuperscript{2}The Hong Kong University of Science and Technology (Guangzhou)
 \\
 \textsuperscript{3}City University of Hong Kong
 \\
}

\newcommand{\syndec}{\textsc{SynDec}}

\begin{document}
	
	\maketitle
	
	\begin{abstract}
		
        Large Language Models~(LLMs) are emerging as dominant forces for textual style transfer. 
        However, for arbitrary style transfer, LLMs face two key challenges: (1) considerable reliance on manually-constructed prompts and (2) rigid stylistic biases inherent in LLMs. 
        In this paper, we propose a novel  \emph{\textbf{Syn}thesize-then-\textbf{Dec}ode} (\textsc{SynDec}) approach, which automatically synthesizes high-quality prompts and amplifies their roles during decoding process. 
        Specifically, our approach synthesizes prompts by selecting representative few-shot samples, conducting a four-dimensional style analysis, and reranking the candidates. At LLM decoding stage, the TST effect is amplified by maximizing the contrast in output probabilities between scenarios with and without the synthesized prompt, as well as between prompts and negative samples.
        We conduct extensive experiments and the results show that \textsc{SynDec} outperforms existing state-of-the-art LLM-based methods on five out of six benchmarks (e.g., achieving up to a 9\% increase in accuracy for modern-to-Elizabethan English transfer). Detailed ablation studies further validate the effectiveness of \textsc{SynDec}.

	\end{abstract}
	
	\section{Introduction}
    In natural language processing (NLP), the task of textual style transfer (TST) holds significant importance. It entails the automatic transformation of text from one stylistic form to another while preserving original content.  TST can enhance readability by transforming complex texts, such as legal documents in Shakespearean style, into plain language \citep{xu-etal-2012-paraphrasing}. In social media management and customer service, TST can adjust the sentiment of responses with  desired tones, making them more polite, empathetic, or assertive \citep{wu_mask_2019}.
    
    Large language models (LLMs), thanks to their advanced text comprehension and contextual learning abilities, can efficiently handle TST tasks with few-shot or even zero-shot learning. Recent studies utilizing LLMs for TST have achieved remarkable success \citep{reif2021recipe, suzgun2022prompt, han-etal-2024-disentangled, lai-etal-2024-style}.
    Despite their effectiveness, LLM-based TST methods still face challenges in the automated construction of prompts and mitigating the rigid stylistic biases of LLMs.

    \textit{\textbf{Prompt construction}}: Current attempts at LLM-based TST methods achieve a large reliance on manually crafted prompts \citep{reif2021recipe, suzgun2022prompt, lai-etal-2024-style}. Creating these manual prompts generally involves analyzing the target text's style, identifying key linguistic features(e.g., vocabulary, syntax, etc.), selecting typical samples, and then using these insights to construct customized prompts for LLMs. This approach demands significant human labor and time.

    
    \textit{\textbf{LLM's Inherent Stylistic Bias}}: LLMs often struggle to fully capture the intended context, leading to generated text that may be inconsistent with the original or contain hallucinations \citep{shi-etal-2024-trusting, chuang2024dola}. In TST tasks, this issue manifests as an inherent stylistic bias, also known as inherent style results \cite{reynolds2021prompt}, where LLMs rely on their prior knowledge rather than the given style transfer instructions or few-shots. This bias can lead to undesired style shifts during the transformation into the target style.

    Motivated by these challenges, we raise the following two research questions:
    
    \noindent\textbf{Q1:} \textit{How can high-quality prompts be automatically synthesized for LLMs in arbitrary TST tasks?}

    \noindent\textbf{Q2:} \textit{After synthesizing prompts, how can LLMs be guided to prioritize prompt content over internal prior knowledge, thereby mitigating stylistic bias?}

   To answer the above questions, we introduce a novel approach called \emph{\textbf{Syn}thesize-then-\textbf{Dec}ode} (\textsc{SynDec}), designed to automatically generate high-quality prompts and enhance their effectiveness in guiding textual style transfer during model decoding.
    \textbf{(1)} \textbf{Synthesizing Stage:} \textsc{SynDec} synthesizes prompts through three key steps: semantic-structural joint sampling, pattern analysis, and few-shots reranking. First, all samples are embedded into a joint semantic-structural space, where clustering algorithms are applied to select those with strong stylistic representation as few-shot samples. Next, an LLM analyzes the style transfer characteristics of these few-shots at four levels: lexis, syntax, tone, and semantics. The few-shots and their corresponding analyses are then combined into analysis chains. Finally, these chains are reranked based on their similarity to the input, and the final prompt is synthesized accordingly.
    \textbf{(2)} \textbf{Decoding Stage:} \textsc{SynDec} ensures that the language model fully attends to the contextual prompts during decoding. Specifically, it amplifies the output probability differences between outputs with and without prompts, as well as between prompts and negative samples, thereby enhancing the guiding effect of prompts on the generated text.

    Experimental results show that our method demonstrates outstanding performances on five public benchmarks and our custom multi-style dataset, surpassing  state-of-the-art (SOTA) LLM-based methods on five out of six benchmarks, e.g., up to an 9\% increase in accuracy for modern-to-Elizabethan English transfer. We further conduct ablation studies to thoroughly investigate the components that influence \textsc{SynDec}'s performance.
    
    In summary, our contributions are as follows:
        	\vspace{-2mm}
    \begin{enumerate}
\item \textbf{A novel synthesize-then-decode approach}, which harnesses LLMs for efficient and effective arbitrary textual style transfer. 
This approach serves two key purposes: \textbf{(1)} significantly reducing the labor-intensive process of prompt engineering and \textbf{(2)} mitigating inherent stylistic biases in LLMs.
\vspace{-2mm}
\item \textbf{An accompanying sampling strategy}, which  considers semantic and structural variations simultaneously to adapt to the arbitrary TST task, ensuring more representative few-shot samples. Its effectiveness is rigorously validated through ablation experiments.\vspace{-2mm}
\item \textbf{Comprehensive baseline comparisons}, encompassing three baseline methods and six benchmark datasets, show that our approach outperforms current SOTA methods on five out of six benchmarks, achieving impressive accuracies of 97\% in sentiment transfer and 99\% in modern English conversion. {\syndec} also achieves leading performance in expert evaluations.\vspace{-2mm}
\item \textbf{A multi-style transfer dataset}, which encompasses two complex multi-style transformation scenarios, serving as a complementary baseline for evaluating new approaches in complex arbitrary TST tasks.
    \end{enumerate}

\section{Related Work}
\textbf{LLM-Based Textual Style Transfer.}
Recent advancements in the field of TST have showcased promising results through the integration of LLMs as the foundation for various methods \citep{reif2021recipe, suzgun2022prompt, roy2023conversation, ostheimer2023text, han-etal-2024-disentangled, lai-etal-2024-style}.
These methods leverage the assistance of a vast knowledge repository to transcend the constraints of parallel data, significantly expanding the scope of achieving arbitrary textual style transfer.

\begin{figure*}[htbp]
	\centering
	 \includegraphics[width=0.95\textwidth]{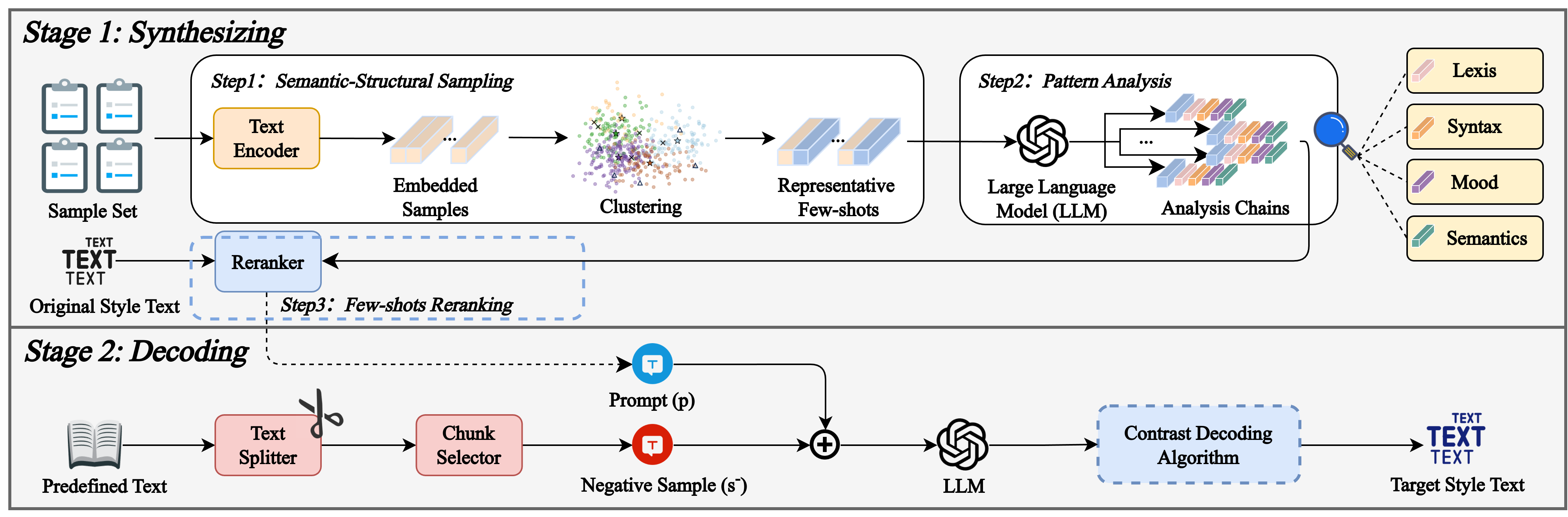}
        \caption{The overview of Synthesize-then-Decode approach.}
	\label{spa}
\end{figure*}

\noindent\textbf{Prompt Construction for LLMs.}
For prompt engineering, writing or constructing prompt words is of critical importance. The in-context learning ability of LLMs heavily relies on the quality of prompts \citep{liu2021pretrain, dong2023survey, wang2023label}. 
Consequently, work on better constructing or designing prompts is being progressively undertaken. To optimize the construction of the prompt, \cite{shin2020autoprompt} shows that the prompts could be auto-refined by using a gradient-guided search. \cite{yang2023large} presents a method that uses LLMs for iterative optimization. Tasks are described in natural language. LLMs create and refine solutions iteratively, enhancing the prompt for future iterations.
However, prompt construction in LLM-based TST still relies on manual methods and has yet to be automated.

\noindent\textbf{Knowledge Conflicts for LLMs.}
Knowledge conflict is a major cause of LLM hallucinations, particularly when contextual knowledge contradicts the model's prior knowledge, leading to insufficient attention to context \citep{longpre-etal-2021-entity, tan-etal-2024-blinded}. Contrastive decoding has shown promise in improving LLM focus on context and mitigating knowledge conflicts \citep{shi-etal-2024-trusting, chuang2024dola}.
Leveraging its capability of reducing knowledge conflicts, we adopt a contrastive learning approach using synthetic prompts to reduce the prior stylistic knowledge of LLMs.

\section{Methodology}
\subsection{Problem Formulation}
We first give the basic notations in this work. We use $\mathcal{T}$ to denote the set of all texts, and $x\in \mathcal{T}$ denotes an instance text. $\mathcal{S}$ denotes the space of all possible text styles, with $s \in \mathcal{S}$ representing a particular style. 
In particular, we denote a text $x$ in style $s$ as a pair $\langle x,s\rangle$. 

\noindent\textbf{Definition 1} (\textit{LLM-based arbitrary textual style transfer}).  
Given an LLM $\ell$, an input text $x$ with an original style $s_1$, a target style $s_2$, and a prompt $p$ that provides an instruction for performing the style transfer, the LLM-based arbitrary textual style transfer task is defined as:
\begin{align}
\langle\hat{x}:\hat{s}_1\rangle = \ell(\langle x, s_1\rangle, s_2, p) \label{1}
\end{align}
where $\hat{x}$ and $\hat{s}_1$ represent the generated text and its corresponding style after the transfer, respectively; $s_1, \hat{s}_1, s_2 \in \mathcal{S}$, and $x, \hat{x}, p \in \mathcal{T}$.

\noindent\textbf{Definition 2} (\textit{Naive decoding in LLM}).
Given an LLM $\ell$, an input $x$ and a prompt $p$, LLM's response can be autoregressive generated from the probability distribution conditioned on $x$ and $p$:
    \begin{align}
    y_t &\sim \tilde{p}_{\theta}(y_t \mid y_{<t}, x, p) \label{2}
    \end{align}
where $t$ represents the time step (generation step), $y_{<t}$ refers to all tokens generated before the current time step $t$, and $y_{t}$ denotes the token generated at the current time step $t$. Here, $y_t, y_{<t}, x, p \in \mathcal{T}$.

\subsection{Synthesize-then-Decode Approach}
 As shown in Figure \ref{spa}, the proposed \emph{\textbf{Syn}thesize-then-\textbf{Dec}ode} (\textsc{SynDec}) approach consists of two stages: synthesizing and decoding. In the synthesizing stage, we sample few-shots, extract their style transfer patterns, and rerank them to construct the final prompt (Section 3.3). In the decoding stage, we introduce negative samples and adjust the model's output probabilities to align with the prompt (Section 3.4), improving the effectiveness of TST.
 In the following sections, we elaborate on the synthesizing and decoding in detail.

\subsection{Synthesizing Stage}
Few-shot learning enables LLMs to quickly adapt to new tasks with only a small number of demonstrations (few-shots) without requiring specialized training \cite{NEURIPS2020_1457c0d6, min-etal-2022-rethinking}. 
To address \textbf{Q1}, we propose an automated TST prompt synthesis pipeline based on few-shot learning. This synthesis process unfolds in three sequential steps: \textit{semantic-structural sampling}, \textit{pattern analysis}, and \textit{few-shots reranking}.

\begin{figure*}[htbp]
	\centering
	 \includegraphics[width=1\textwidth]{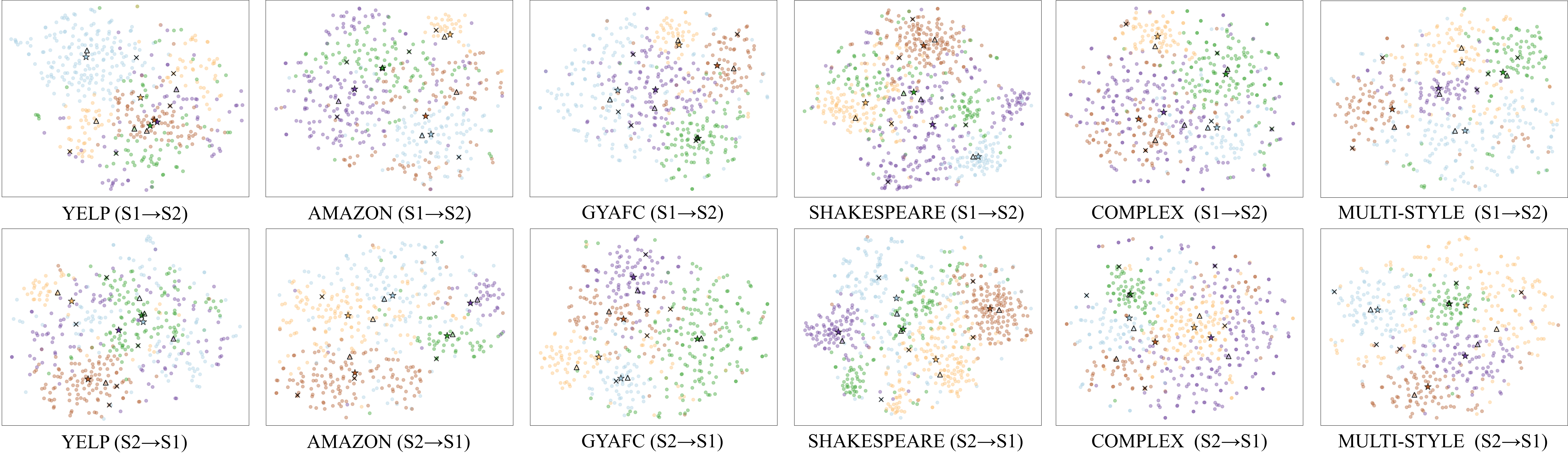
     }
        \caption{Clustering results ($K$=5) on six TST datasets. Stars indicate few-shots selected by {\syndec}, crosses represent those obtained via random sampling, and triangles denote manually annotated representative samples.}
	\label{clustering}
\end{figure*}

\subsubsection{Semantic-Structural Sampling}

In existing LLM-based TST methods \citep{suzgun2022prompt, 10.1609/aaai.v38i17.29832}, manually selected few-shots are typically representative in both semantic and structural dimensions. These few-shots play a crucial role in guiding the model to perform accurate style transfer while maintaining the integrity of the original content. To automate this selection process, we propose a novel sampling strategy that identifies representative few-shots by embedding the samples into a joint semantic-structural space and applying unsupervised clustering. Figure \ref{clustering} shows that {\syndec}'s few-shots are closer to manual ones than those from random sampling.

\textbf{Embedding:}  
We employ a sentence embedding method based on \textbf{Directed Graph Convolutional Networks (DGCNs)} inspired by \citep{vashishth-etal-2019-incorporating, kumar-etal-2020-syntax} to integrate both syntactic dependencies and semantic features. Specifically, for a given sentence pair \(q_i\), we first parse its dependency structure to construct a directed graph
\[
G_{q_i} = \bigl(V_{q_i}, E_{q_i}\bigr),
\]
where \(V_{q_i} = \{w_1, w_2, \dots, w_n\}\) is the set of word nodes, and \(E_{q_i}\) contains labeled directed edges in the form \((w_u, w_v, r_{uv})\). Here, \(r_{uv}\) is the dependency relation label between \(w_u\) and \(w_v\). We define the neighbor set for each node \(w_i\) as \(\mathcal{N}(i)\).
To capture syntactic and semantic information, we stack \(k\) layers of graph convolutions. Let \(h_i^{(\ell)}\) denote the representation of node \(i\) at the \(\ell\)-th layer. Then the update rule for the \((\ell+1)\)-th layer is given by:
\[
h_i^{(\ell+1)} 
= \phi\!\Biggl(\sum_{j \in \mathcal{N}(i)} \alpha_{ij}^{(\ell)} \Bigl(W_{r_{ij}}^{(\ell)}\,h_j^{(\ell)} + b_{r_{ij}}^{(\ell)}\Bigr)\Biggr),
\]
where \(\phi(\cdot)\) is the activation function ReLU \citep{agarap2019deeplearningusingrectified}, and \(W_{r_{ij}}^{(\ell)}, b_{r_{ij}}^{(\ell)}\) are trainable parameters associated with the dependency relation label \(r_{ij}\). The coefficient \(\alpha_{ij}^{(\ell)}\) is used to weight the contribution from each neighbor and is computed as follows:
\[
\alpha_{ij}^{(\ell)} 
= \sigma\!\Bigl(\tilde{W}_{r_{ij}}^{(\ell)}\,h_i^{(\ell)} + \tilde{b}_{r_{ij}}^{(\ell)}\Bigr),
\]
where \(\sigma(\cdot)\) denotes the sigmoid function, and \(\tilde{W}_{r_{ij}}^{(\ell)}\) and \(\tilde{b}_{r_{ij}}^{(\ell)}\) are also trainable parameters.
After going through all \(k\) layers of graph convolution, we aggregate the final node representations \(\{h_j^{(k)}\}\) by applying average pooling:
\[
x_i = \frac{1}{|V_{q_i}|} \sum_{w_j \in V_{q_i}} h_j^{(k)},
\]
which yields the sentence-level embedding. This procedure effectively incorporates structural information from the dependency graph into the final sentence vector representation, yielding a richer semantic-structural joint space.

\textbf{Clustering:}  
To partition the vector representations $\{x_n\}_{n=1}^N$ in the semantic-structural joint space, we employ a modified $k$-means++ \cite{10.5555/1283383.1283494} clustering algorithm. 
Clustering algorithm \ref{alg:clustering} begins by randomly selecting the first center $\mu_1$. Each subsequent center $\mu_k$ is chosen with probability proportional to the squared distance from the nearest existing center, ensuring well-separated initial clusters. After $K$ centers are initialized, data points are assigned to the nearest one, and centroids are iteratively updated until convergence.  
The outputs are sentence pairs \( Q^* = ({pair}_1, {pair}_2, \dots, {pair}_n) \), obtained by mapping the cluster centroids $\{\mu_k\}_{k=1}^K$ back to natural language.

\begin{algorithm}[t]
\caption{Clustering Algorithm}\label{alg:clustering}
\begin{algorithmic}[1]
\renewcommand{\algorithmicrequire}{\textbf{Input:}}
\renewcommand{\algorithmicensure}{\textbf{Output:}}
\REQUIRE \textit{Set of vector representations $\{x_n\}_{n=1}^N$, number of clusters $K$}
\ENSURE \textit{Set of representative pairs $Q^*$}
\STATE $n \gets$ \texttt{RandomInteger}(1, $N$)
\STATE $\mu_1 \gets x_n$ \hfill\textit{\hil{// Set the first cluster center $\mu_1$}}
\STATE \textit{Initialize distance array $D \gets \infty$ of size $N$ }
\FOR{$k \gets 2$ \ldots $K$}
    \FOR{$n \gets 1$ \ldots $N$}
        \STATE $D[n] \gets \min(D[n], \| x_n - \mu_{k-1} \|_2)$ 
    \ENDFOR
    \STATE $q \gets D^2 / \sum D^2$ \hfill\textit{\hil{// Compute probability distribution}}
    \STATE $n \gets$ \texttt{Discrete}($p$)
    \STATE $\mu_k \gets x_n$ \hfill\textit{\hil{// Set $x_n$ as the new center $\mu_k$}}
\ENDFOR
\STATE \textit{De-embed cluster means $\{\mu_k\}_{k=1}^K$ into $Q^*$}
\RETURN $Q^*$
\end{algorithmic}
\end{algorithm}

\subsubsection{Pattern Analysis}
After selecting representative few-shots, the next step is to analyze their stylistic transfer patterns. This analysis aims to provide clearer guidelines for the LLM to follow when performing textual style transfer. We examined these few-shots across four key style transfer dimensions: \textit{lexis}, \textit{syntax}, \textit{tone}, and \textit{semantics}, using LLMs with predefined descriptive prompts to facilitate context learning.
In Example 1, descriptive prompts 1, 2, 3, and 4 describe the style transfer characteristics of the few-shots from the four dimensions, respectively.
\begin{tcolorbox}[
  colback=blue!5!white,              
  toprule=0mm,
  bottomrule=0mm,
  leftrule=0.7mm,
  rightrule=0mm,
  titlerule=0.2mm,
  coltitle=black,                    
  fonttitle=\scshape,               
  colbacktitle=blue!5!white,         
  title=\textbf{Example 1}: Descriptive Prompts, 
  sharp corners=all,                 
  top=1mm, bottom=1mm,               
  left=1mm, right=1mm                
]
\small 
\textbf{\scshape{Prompt 1:}} 
\textcolor[RGB]{19,165,71}{\texttt{[onebest]}}:\textit{"Analyze the lexical variations between the following sentence pairs in terms of word choice, vocabulary, and stylistic expression."}, \textcolor{red}{\texttt{[type]}}: \textit{Lexis}

\textbf{\scshape{Prompt 2:}} 
\textcolor[RGB]{19,165,71}{\texttt{[onebest]}}:\textit{"Examine and compare the syntactic structures of these style transfer sentence pairs, focusing on sentence construction, grammatical patterns, and syntax differences."}, \textcolor{red}{\texttt{[type]}}: \textit{Syntax}

\textbf{\scshape{Prompt 3:}} 
\textcolor[RGB]{19,165,71}{\texttt{[onebest]}}:\textit{"Evaluate the tone of these texts by comparing their mood, emotional cues, and overall attitude towards the subject matter."}, \textcolor{red}{\texttt{[type]}}: \textit{Mood}

\textbf{\scshape{Prompt 4:}} 
\textcolor[RGB]{19,165,71}{\texttt{[onebest]}}:\textit{"Analyze the semantic shifts between these sentence pairs, identifying differences in meaning, context, and interpretation."}, \textcolor{red}{\texttt{[type]}}: \textit{Semantics}
\end{tcolorbox}

\textbf{Analysis Chain:} 
Next, we organize the analysis of the four dimensions into an analysis chain format. In Example 2, each analysis chain corresponds to a representative few-shot sample, along with the lexis analysis, syntax analysis, mood analysis, and semantics analysis for that sample.
\begin{tcolorbox}[
  colback=blue!5!white,              
  toprule=0mm,
  bottomrule=0mm,
  leftrule=0.7mm,
  rightrule=0mm,
  titlerule=0.2mm,
  coltitle=black,                    
  fonttitle=\scshape,               
  colbacktitle=blue!5!white,         
  title=\textbf{Example 2}: Analysis Chain, 
  sharp corners=all,                 
  top=1mm, bottom=1mm,               
  left=1mm, right=1mm                
]
\small 
\textbf{\scshape{Chain 1:}} 
\textcolor[RGB]{19,165,71}{\texttt{[Sample 1]}} \textcolor[RGB]{19,165,71}{\texttt{[Analysis 1-1]}} \textcolor{red}{\texttt{[type]}}: \textit{Lexis}, \textcolor[RGB]{19,165,71}{\texttt{[Analysis 1-2]}} \textcolor{red}{\texttt{[type]}}: \textit{Syntax}, \textcolor[RGB]{19,165,71}{\texttt{[Analysis 1-3]}} \textcolor{red}{\texttt{[type]}}: \textit{Mood}, \textcolor[RGB]{19,165,71}{\texttt{[Analysis 1-4]}} \textcolor{red}{\texttt{[type]}}: \textit{Semantics}

\textbf{\scshape{Chain 2:}} 
\textcolor[RGB]{19,165,71}{\texttt{[Sample 2]}} \textcolor[RGB]{19,165,71}{\texttt{[Analysis 2-1]}} \textcolor{red}{\texttt{[type]}}: \textit{Lexis}, \textcolor[RGB]{19,165,71}{\texttt{[Analysis 2-2]}} \textcolor{red}{\texttt{[type]}}: \textit{Syntax}, \textcolor[RGB]{19,165,71}{\texttt{[Analysis 2-3]}} \textcolor{red}{\texttt{[type]}}: \textit{Mood}, \textcolor[RGB]{19,165,71}{\texttt{[Analysis 2-4]}} \textcolor{red}{\texttt{[type]}}: \textit{Semantics}

\end{tcolorbox}

\subsubsection{Few-shots Reranking}

Existing studies \citep{kumar-talukdar-2021-reordering, guo-etal-2024-sample} have shown that the ordering of few-shots significantly impacts the effectiveness of prompt construction. By placing samples with higher similarity to the input sentence \( x \) at the front, better contextual alignment can be achieved. To this end, we embed the input sentence \( x \) together with the few-shots \( Q^{*} = (pair_1, pair_2, \ldots, pair_n) \) into a joint semantic-structural space, obtaining their vector representations \( \hat{x} \) and \( \{\hat{q}_i\}_{i=1}^n \), respectively. We then calculate the cosine similarity between \( \hat{x} \) and each \( \hat{q}_i \), and sort all samples in descending order: \( \text{Rank}(Q^{*}) = \mathrm{Sort}_{desc}(\{ \cos(\hat{x}, \hat{q}_1), \ldots, \cos(\hat{x}, \hat{q}_n) \}) \). The reordered samples and their corresponding analysis chains are subsequently incorporated into the system prompt to generate the final prompt \( p \).



\subsection{Decoding Stage}  
To address \textbf{Q2}, we employ a contrastive decoding strategy, which enhances LLM output probabilities through contrastive learning \cite{shi-etal-2024-trusting, zhao-etal-2024-enhancing}. By comparing probabilities with and without the prompt, as well as with positive and negative samples, we aim to strengthen the effect of prompt \( p \) and reduce stylistic bias.  
The decoding process of {\syndec} is formulated as:  
\begin{align}  
y_t &\sim \tilde{p}_{\theta}(y_t \mid p, s^{-}, x, y_{<t}) \\  
&\propto p_{\theta}(y_t \mid p, x, y_{<t}) \left( \frac{p_{\theta}(y_t \mid p, x, y_{<t})}{p_{\theta}(y_t \mid x, y_{<t})} \right)^\alpha \nonumber \\  
&\quad \times \left( \frac{p_{\theta}(y_t \mid p, x, y_{<t})}{p_{\theta}(y_t \mid s^{-}, x, y_{<t})} \right)^\beta  \label{5}
\end{align}  

The first term, \( p_{\theta}(y_t \mid p, x, y_{<t}) \), represents the model's prediction given prompt \( p \), while \( p_{\theta}(y_t \mid x, y_{<t}) \) reflects its response to \( x \) based solely on internal parameters. To enhance prompt adherence and reduce the LLM’s inherent stylistic bias, we amplify the probability difference between them with parameter \( \alpha \).  
\( s^{-} \) denotes a negative sample unrelated to TST, which may hinder style transfer effectiveness. We use a pre-written long text as the negative sample, detailed in the next section.  
Finally, the term \( \left( \frac{p_{\theta}(y_t \mid p^+, x, y_{<t})}{p_{\theta}(y_t \mid s^-, x, y_{<t})} \right)^\beta \) adjusts predictions based on the contrast between prompt \( p \) (positive sample) and the negative sample \( s^- \), reducing the likelihood of generating tokens aligned with \( s^- \), controlled by parameter \( \beta \).  
We normalize Eq.\ref{5} to obtain a valid probability distribution:
\begin{align}
y_t \sim \text{softmax} [
    &(1 + \alpha + \beta) \log p_{\theta}(y_t \mid p, x, y_{<t}) \nonumber \\
    &- \alpha \log p_{\theta}(y_t \mid x, y_{<t}) \nonumber \\
    &- \beta \log p_{\theta}(y_t \mid s^{-}, x, y_{<t}) ].
\end{align}

In essence, {\syndec} adjusts the model's predictions based on both \( p \) and \( s^- \), enabling controlled style shifts while mitigating the LLM's inherent stylistic bias and misleading information. If \( \alpha = 0 \) and \( \beta = 0 \), the model defaults to naive decoding, as shown in Eq.\ref{2}.

\subsubsection{Construction of Negative Sample \texorpdfstring{$s^{-}$}{s-}}
To construct the negative sample \( s^{-} \), we carefully selected text from the irrelevant context constructed in \cite{zhao-etal-2024-enhancing}, ensuring minimal relevance. Specifically, we pre-segmented the irrelevant context into fixed-size text chunks with \textit{langchain's recursive text splitter}\footnote{https://github.com/langchain-ai/langchain}. During each decoding process, the segment with the lowest similarity to the prompt in the semantic-structural joint space was selected as \( s^{-} \) for contrastive decoding.

\subsubsection{\texorpdfstring{Trade-off Parameters $\alpha$ and $\beta$}{Trade-off Parameters alpha and beta}}

Bayesian optimization was employed to fine-tune the trade-off parameters $\alpha$ and $\beta$ for maximizing validation performance. 
Optimization was conducted via \textit{Optuna}\footnote{https://github.com/optuna/optuna}, leveraging Gaussian processes to model the search space. Starting with initial values of $\alpha = \beta =5$, the algorithm iteratively refined these parameters, systematically exploring and updating the optimization landscape.  


\section{Experiments}

\subsection{Experimental Setup}
\textbf{Tasks and Datasets.} We evaluate style transfer capability on the following benchmarks.
\begin{itemize}
	\item Sentiment transfer. For this task, we use the Yelp polarity dataset \citep{zhang2015character} and the Amazon reviews dataset \citep{li-etal-2018-delete} as the benchmarks for sentiment transformation.
	\item Formality transfer. We utilize the Formality Corpus (GYAFC) dataset, as described in DeleteAndRetrieve \citep{li-etal-2018-delete}, to demonstrate variations in formality.
	\item Elizabethan-to-modern English transfer. To evaluate the capability of transferring Elizabethan English to modern English, we employ the SHAKESPEARE dataset \citep{xu-etal-2012-paraphrasing}.
	\item Complexity transfer. We use the COMPLEX dataset \citep{xu-etal-2016-optimizing}, which consists of parallel corpora containing original sentences and their corresponding simplified versions.
	\item Multi-style transfer. We evaluate with a custom multi-style dataset that transfers negative Elizabethan English to positive modern English. The dataset construction process is detailed in the supplementary material.
\end{itemize}
To mitigate the effects of spaces or unusual characters on TST, we employed the cleaning tools from PromptAndRerank \citep{suzgun2022prompt} to enhance the quality of these datasets.

\vskip 2mm
\noindent\textbf{Baseline.}
We compare {\syndec} against the following representative baselines:
(1) \textbf{LLaMA-3} \citep{grattafiori2024llama3herdmodels}: the pretrained LLaMA-3 model without any additional fine-tuning, serving as a vanilla baseline.
(2) \textbf{PEGF} \citep{liu-etal-2024-step}: a method that steers LLMs to perform style transfer by editing only a minimal portion of the text within a localized editing region.
(3) \textbf{APR} \citep{10.1609/aaai.v38i17.29832}: a prompt-routing approach that selects the most suitable prompt based on input characteristics to enhance style transformation.

\begin{table*}[!t]

	\centering
	\footnotesize
		\setlength{\tabcolsep}{6pt}
        \scalebox{0.94}{
\begin{tabular}{llcccccccc}
    \toprule
    \multirow{2}{*}{Dataset} & \multirow{2}{*}{Model} & 
    \multicolumn{4}{c}{\boldmath{${S_1}\rightarrow{S_2}$}} & \multicolumn{4}{c}{\boldmath{${S_2}\rightarrow{S_1}$}} \cr
    \cmidrule(r){3-6} \cmidrule(r){7-10}
    & & Acc$\uparrow$ & r-sBLEU$\uparrow$ & s-sBLEU$\uparrow$ & PPL$\downarrow$ & Acc$\uparrow$ & r-sBLEU$\uparrow$ & s-sBLEU$\uparrow$ & PPL$\downarrow$ \\
    
    \midrule
    & LLaMA-3 & 0.93 & 11.7 & 19.2 & 133 & 0.85 & \underline{13.1} & 22.9 & \textbf{82} \\
    \multirow{2}{*}{YELP}& PEGF & 0.95 & 10.2 & 18.0 & 120 & 0.83 & 12.9 & 21.4 & 107 \\
     & APR & \underline{0.97} & \underline{27.1} & \underline{39.4} & \textbf{101} & \textbf{0.87} & 12.0 & \underline{25.3} & 105 \\ 
    & {\syndec} (Ours) & \textbf{0.97} & \textbf{30.9} & \textbf{51.3} & \underline{109} & \underline{0.86} & \textbf{15.3} & \textbf{28.2} & \underline{95} \\ 
    
    \midrule
    & LLaMA-3 & 0.85 & 25.4 & 39.8 & 160 & 0.70 & \underline{19.9} & \underline{31.5} & 112 \\
    \multirow{2}{*}{AMAZON}& PEGF & \underline{0.86} & \textbf{33.2} & \textbf{41.7} & \textbf{98} & \textbf{0.78} & 15.7 & 22.4 & \underline{71} \\
     & APR & 0.77 & 20.9 & 31.2 & 150 & 0.61 & 13.2 & 19.9 & \textbf{50} \\ 
    & {\syndec} (Ours)& \textbf{0.91} & \underline{28.1} & \underline{40.6} & \underline{132} & \underline{0.73} & \textbf{20.5} & \textbf{45.1} & 109 
    \\
    
    \midrule
    & LLaMA-3 & 0.81 & 15.4 & 33.7 & \underline{96} & 0.75 & 20.4 & \underline{41.9} & 124 \\
    \multirow{2}{*}{GYAFC}& PEGF & \underline{0.88}	&\textbf{23.1}&	\textbf{40.1}&	132 & 0.81 & 7.1 & 5.2 & \underline{63} \\ 
     & APR & 0.78&	8.9&	15.3&	103 & \underline{0.85} & \textbf{36.4} & \textbf{49.6} & 68 \\
    & {\syndec} (Ours)& \textbf{0.89} & \underline{18.5} & \underline{34.1} & \textbf{91} & \textbf{0.92} & \underline{22.0} & 38.2 & \textbf{61} \\ 
    
    \midrule
    & LLaMA-3 & 0.87 & 8.7 & 14.2 & 124& 0.71 & 15.3 & 28.4 & \underline{78} \\
    \multirow{2}{*}{SHAKESPEARE}& PEGF & 0.87	&13.1&	19.9	&\underline{92}  & 0.70 & 8.5 & 13.1 & 98 \\ 
     & APR & \underline{0.93}	&\underline{14.2}	&\underline{22.1}	&161 & \underline{0.74} &\underline{17.1} & \underline{40.6} & 97 \\
    & {\syndec} (Ours)& \textbf{0.99} & \textbf{17.7} & \textbf{23.0} & \textbf{80} & \textbf{0.83} & \textbf{44.6} & \textbf{48.2} & \textbf{64} \\ 
    
    \midrule
    & LLaMA-3 & 0.76 & 15.7 & 30.8 & 76 & 0.51 & 39.5 & 51.2 & 101 \\
    \multirow{2}{*}{COMPLEX}& PEGF &\underline{0.85} & \underline{21.5} & \underline{42.4} & 106 & \underline{0.59} & 33.5 & 50.4 & \underline{82} \\ 
     & APR & 0.82 & 20.9 & 38.0 & \underline{75} & 0.53 &\underline{42.1} & \underline{61.0} & \textbf{79} \\
    & {\syndec} (Ours)& \textbf{0.88} & \textbf{22.4} & \textbf{45.9} & \textbf{66} & \textbf{0.60} & \textbf{42.4} & \textbf{61.9} & 124 \\ 
    
    \midrule
    & LLaMA-3 & 0.54 & \underline{15.6} & \underline{30.9} & 137 & 0.60 & 14.9 & 28.1 & 136 \\
    \multirow{2}{*}{MULTI-STYLE}& PEGF & 0.63	&12.3	&21.5	&176& \underline{0.69} & 12.2 & 22.4 & \underline{88} \\ 
     & APR & \underline{0.71}	&8.0	&11.9	&\underline{99}& 0.62 &  \underline{15.3} &  \underline{28.5} & 95 \\
    & {\syndec} (Ours)& \textbf{0.81} & \textbf{24.2} & \textbf{45.0} & \textbf{54} & \textbf{0.76} & \textbf{22.6} & \textbf{48.2} & \textbf{75} \\ 
    \bottomrule
\end{tabular}

}
        \caption{Comparison of {\syndec} with baseline models across full benchmarks. ${S_1}\rightarrow{S_2}$ represents positive text style transfer, while ${S_2}\rightarrow{S_1}$ indicates the reverse. Bold values represent the highest values, and underlined values represent the second-highest values.}
	\label{table2}
\end{table*}

\vskip 2mm
\noindent\textbf{Evaluation metrics.}
We focus on three dimensions to evaluate the model's performance on textual style transfer.
\begin{itemize}
	\item \textit{Style transfer accuracy.} We follow the prior work \citep{mir-etal-2019-evaluating} to measure style transfer accuracy by employing a sentence-level style classifier. We fine-tune RoBERTa-Large \citep{liu2019roberta} on the training dataset for each task as the sentence-level style binary classifier.
	
	\item \textit{Content preservation.} We use SacreBLEU \citep{post-2018-call} to measure the ability of content preservation. By calculating reference-BLEU (r-sBLEU) and self-sBLEU (s-sBLEU), we measure the preservation of the replicated source in generated sentences.
	
	\item \textit{Fluency.} Fluency in TST is typically measured using the perplexity of a language model. We use GPT2-Large \citep{Radford2019LanguageMA} as the language model for perplexity (PPL) evaluation.

\end{itemize}
\noindent\textbf{Expert Evaluation.} 
We conducted an expert evaluation to assess the effectiveness of TST, focusing on three widely used key aspects: \textit{style transfer accuracy}, \textit{content preservation}, and \textit{fluency}. Each aspect was rated on a 1–5 Likert scale, with 5 indicating the highest performance. The evaluation covered whole five tasks and included a total of 240 style transfer examples, encompassing outputs from both \textsc{SynDec} and baselines.
Fifteen linguistics graduate students, all trained in stylistic analysis, served as evaluators. Each evaluator spent an average of three hours on the task, and each example was rated by at least three annotators. To assess inter-rater reliability, we used Fleiss' Kappa as the agreement metric among evaluators. The resulting score was 0.921, indicating a high level of agreement and confirming the reliability of the expert judgments.

\noindent\textbf{Implementation.} 
Each dataset was split into training, sample, and test sets in an 8:1:1 ratio. The finetuning process for the binary classifier in style transfer accuracy utilized the AdamW optimizer \cite{loshchilov2017decoupled} with parameters $\beta_1 = 0.9$, $\beta_2 = 0.999$, and $\epsilon = 10^{-8}$. The initial learning rate was set at $2 \times 10^{-5}$, a batch size of 16, and training was carried out over 20 epochs.
All LLM-based methods adopted LLaMA3-70B as the backbone model, using the same hyperparameter settings as reported in their original papers.

\subsection{Results}
\begin{figure*}[t]
    \centering
    \includegraphics[width=1\linewidth]{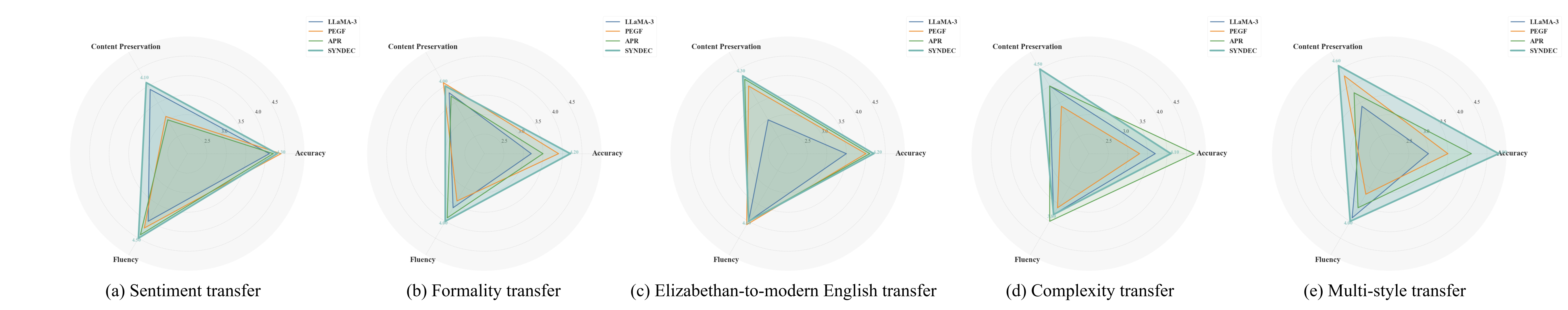}
    \caption{Expert evaluation of all methods on five TST tasks based on style transfer accuracy, content preservation, and fluency, with scores ranging from 0 to 5.}
    \label{pie}
\end{figure*}

\noindent\textbf{Competitive performance of {\syndec} on six benchmarks.} Table \ref{table2} presents a comparative analysis of our approach against other leading LLM-based methods. {\syndec} consistently outperforms competing methods in style transfer accuracy across all benchmarks for the $S_1 \rightarrow S_2$ direction, achieving an impressive 99\% accuracy and a low perplexity score of 80 for the transfer from Elizabethan to modern English. In more challenging tasks, such as multi-style transfer, {\syndec} also proves highly effective, outperforming other methods by 10\% in accuracy, 8.6 in r-sBLEU, 14.1 in s-sBLEU, and 45 in perplexity. For the $S_2 \rightarrow S_1$ transfer, {\syndec} demonstrates strong robustness in style transfer accuracy and significantly surpasses other methods in content preservation, achieving the highest scores on five out of six benchmarks.
\noindent\textbf{Expert evaluation.}
Figure \ref{pie} presents the results of our expert evaluation. The proposed method, {\syndec}, consistently achieved the highest fluency scores across all TST tasks and received top ratings from human evaluators in both style transfer accuracy and content preservation on three out of five TST tasks. We also observed that the APR method demonstrated superior style transfer accuracy on the complexity task, while PEGF exhibited stronger content preservation on the formality task.

\subsection{Analysis}

\noindent\textbf{Ablation Study.}
We conducted ablation experiments on {\syndec}, focusing on the key components involved in prompt synthesis (\textbf{Q1}) and the decoding strategy (\textbf{Q2}). The Yelp and Multi-style datasets were used for evaluation.

Based on the results presented in Tables \ref{table4} and \ref{table5}, we summarize the following key findings:
\textbf{(1)} Among the components of prompt synthesis, sampling and analysis chains are the most critical. Using random sampling or omitting the analysis chain significantly reduces style transfer accuracy, primarily because the model lacks clear guidance during style transformation, leading to lower-quality outputs. In contrast, ablating reranking shows some degradation, but its impact is relatively limited.
\textbf{(2)} Replacing contrastive decoding strategy with naive decoding leads to a substantial drop in performance across all evaluation metrics. This indicates that naive decoding fails to effectively mitigate the inherent stylistic biases of LLMs and cannot fully leverage prompt information for fine-grained style control.
\textbf{(3)} Ablation results show that removing negative sample alignment strongly affects perplexity, while removing prompt alignment causes substantial degradation across all key metrics.

\begin{table}[!t]
	\centering
		\setlength{\tabcolsep}{1.8pt}
		\footnotesize
		\begin{tabular}{lcccc}
			\toprule
		     Methods  & Acc$\uparrow$ &\enspace r-sBLEU$\uparrow$ & \enspace s-sBLEU$\uparrow$ &\enspace PPL$\downarrow$ \\
			\midrule
                                \multicolumn{5}{l}{\textit{Yelp Dataset}} \\
                {\syndec} (Ours)&0.97 & 30.9 & 51.3 & 109\\
                (-) sampling & 0.94 &30.1 &49.5 &134\\
			(-) analysis chains & 0.95 &28.4 &40.1 &98 \\ 
			(-) reranking & 0.96 &18.5 &34.8 &101 \\ 
            \midrule
                    \multicolumn{5}{l}{\textit{Multi-style Dataset}} \\
                {\syndec} (Ours)&0.81 & 24.2 & 45.0 & 54\\
                (-) sampling & 0.76 &23.1 &40.7 &78\\
			(-) analysis chains & 0.78 &18.5 &30.1 &66 \\ 
			(-) reranking & 0.80 &23.5 &42.9 &62 \\ 
			\bottomrule
		\end{tabular}
  	\caption{Ablation study of the prompt synthesis on Yelp and Multi-style datasets.}
	\label{table4}
\end{table}

\begin{table}[!t]
	\centering
	\setlength{\tabcolsep}{1.8pt}
	\footnotesize
	\begin{tabular}{lcccc}
		\toprule
		\textbf{Methods} & Acc$\uparrow$ & r-sBLEU$\uparrow$ & s-sBLEU$\uparrow$ & PPL$\downarrow$ \\
		\midrule
        \multicolumn{5}{l}{\textit{Yelp Dataset}} \\
        ${\syndec}_{\mathit{CD}}$ & 0.97 & 30.9 & 51.3 & 109 \\
        ${\syndec}_{\mathit{ND}}$ & 0.92 & 29.3 & 56.5 & 132 
        \\
        ${\syndec}_{\mathit{w/o\ prompt}}$ & 0.96 & 33.4 & 60.1 & 127 \\
        ${\syndec}_{\mathit{w/o\ neg.\ sample}}$ & 0.97 & 28.5 & 46.9 & 159 \\
        \midrule
        \multicolumn{5}{l}{\textit{Multi-style Dataset}} \\
        ${\syndec}_{\mathit{CD}}$ & 0.81 & 24.2 & 45.0 & 54 \\
        ${\syndec}_{\mathit{ND}}$ & 0.70 & 10.7 & 39.5 & 101 \\
        ${\syndec}_{\mathit{w/o\ prompt}}$ & 0.78 & 15.8 & 34.0 & 71 \\
        ${\syndec}_{\mathit{w/o\ neg.\ sample}}$ & 0.80 & 21.6 & 40.5 & 89 \\
		\bottomrule
	\end{tabular}
	\caption{Ablation study of decoding strategy on Yelp and Multi-style datasets. $\mathit{CD}$ denotes contrastive decoding, and $\mathit{ND}$ denotes naive decoding.}
	\label{table5}
\end{table}



\begin{figure}[h!]
	\centering
	\includegraphics[width=1\linewidth]{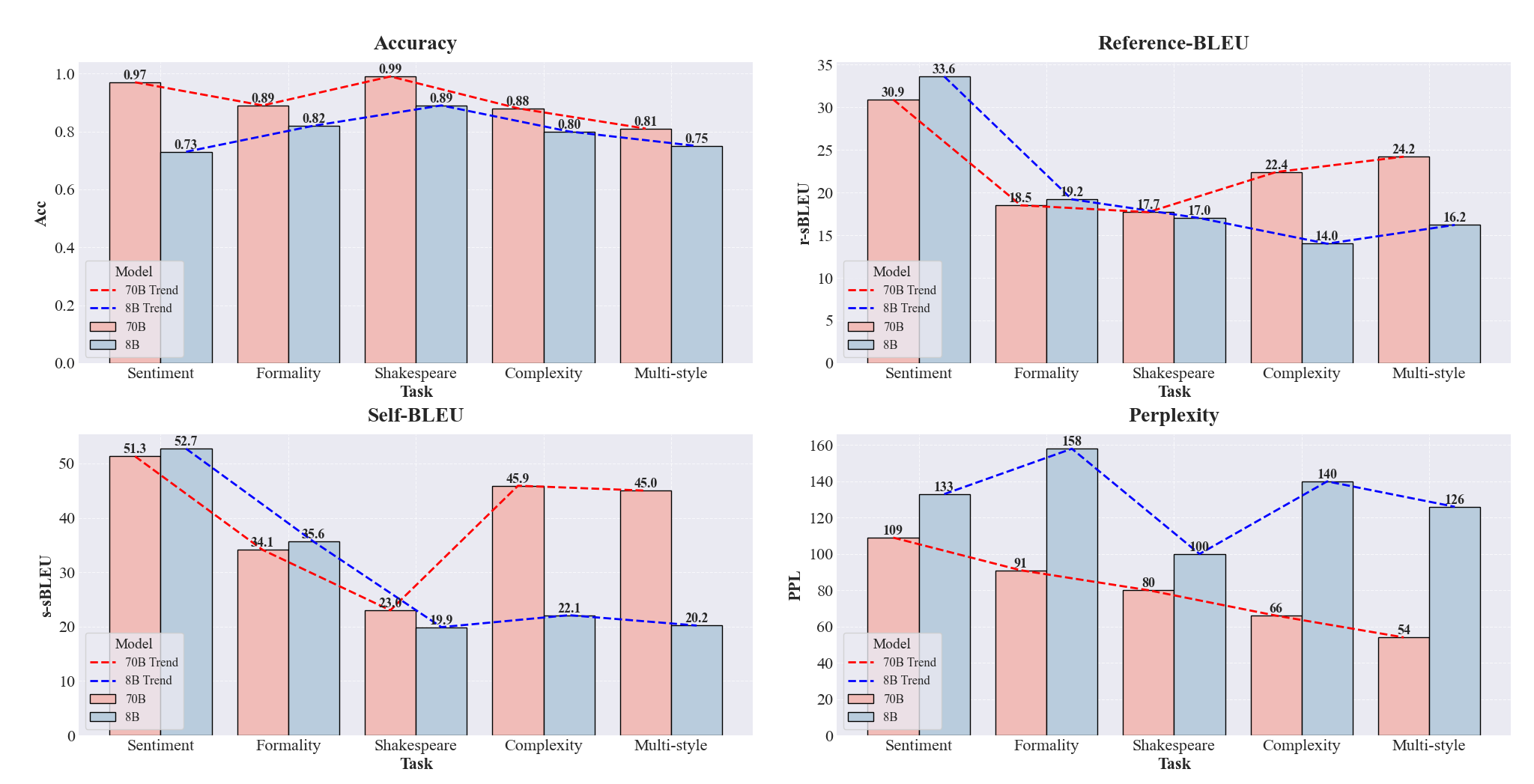}
	\caption{Effect of the LLM's scope of {\syndec} on five TST tasks.}
	\label{scope}
\end{figure}

\noindent\textbf{Scale of LLM.}
We conduct a comprehensive study on the impact of the LLM scale on {\syndec} performance, as shown in Figure \ref{scope}. Our key observations are as follows: \textbf{(1)} As the scale of the LLM decreases, the style transfer accuracy of {\syndec} gradually declines across various TST tasks.
\textbf{(2)} For more complex tasks, such as complexity and multi-style tasks, the negative impact of reducing the LLM scale becomes more pronounced, reflected by a significant drop in content preservation metrics (e.g., s-sBLEU and r-sBLEU) and an increase in perplexity.

\section{Conclusion}

In this paper, we introduce {\syndec}, an innovative technique designed to address the key challenges in LLM-based textual style transfer. By automatically synthesizing high-quality prompts and enhancing their impact during the decoding process, {\syndec} significantly improves style transfer capabilities. Experimental results demonstrate that {\syndec} consistently outperforms existing LLM-based methods, validating its effectiveness.

\newpage
\section*{Limitations}
While \emph{{\syndec}} has reached outstanding performance in experiments, our paper still has the following three limitations:

\begin{enumerate}
    \item Our approach has so far only explored prompt synthesis and decoding strategies for LLMs, with insufficient investigation into the internal mechanisms such as neurons and network layers that influence style within the LLM.
    \item Currently, our {\syndec} approach is limited by its reliance on few-shots. In zero-shot scenarios, where no dataset is available for sampling, our method struggles to finish synthesizing stage.
    \item This study primarily focuses on English. The adaptability to other languages and cultural contexts has not been thoroughly examined. Future research should extend to multilingual settings to explore effective strategies for cross-cultural style transfer.
\end{enumerate}
\clearpage

\bibliography{acl_latex}

\clearpage
\appendix

\newpage

\section{Multi-Style Benchmark Construction}
\begin{table*}
\footnotesize
\setlength{\tabcolsep}{1pt}
\centering
\begin{tabular}{lllc}
\toprule
Dataset & Styles & Example Sentence-Pairs & Test Set Size \\
\midrule
Yelp Restaurant Reviews & Negative & \textit{She was not happy being there.} & \multirow{2}{*}{1000} \\
\citep{zhang2015character} & Positive & \textit{She seemed happy to be there.} & \\
\midrule
Amazon Product Reviews & Negative & \textit{I am actually afraid to open the remaining jars.} & \multirow{2}{*}{1000} \\
\citep{li-etal-2018-delete} & Positive & \textit{I am not afraid to open the remaining jars.} & \\
\midrule
GYAFC Formality Dataset & Informal & \textit{Ask him to go see a doc.} & \multirow{2}{*}{1000} \\
\citep{li-etal-2018-delete} & Formal & \textit{Ask him if you should go see a doctor.} & \\
\midrule
Shakespearean English Dataset & Elizabethan & \textit{Here, sir, a ring she bid me give you, sir.} & \multirow{2}{*}{600} \\
\citep{xu-etal-2012-paraphrasing} & Modern & \textit{Here, sir, this is a ring she asked me to give you.} & \\
\midrule
Complexity Dataset & Complex & \textit{He makes a significant contribution to the world.} & \multirow{2}{*}{600} \\
\citep{xu-etal-2016-optimizing} & Simple & \textit{He contributes to the world greatly.} & \\
\midrule
Multi-Style dataset & Elizabethan, Negative & \textit{Find thou the means, and I'll not find such a man.} & \multirow{2}{*}{600} \\
(\textit{Ours}) & Modern, Positive & \textit{Find out the way, and I'll find the right man.} & \\
\bottomrule
\end{tabular}
\caption{\label{tab:datasets}
The overview of the TST benchmarks.
}
\end{table*}

The current TST task primarily focuses on single-style conversion, where a text is transformed from one style to another.
However, in real-world scenarios, texts often exhibit composite characteristics, showcasing multiple stylistic attributes simultaneously.
For example, a passage may reflect the Shakespearean style while also conveying negative sentiment. Therefore, our goal is to construct a multi-style transfer dataset to address this gap in existing research. 
We begin with the Shakespearean-style dataset \citep{xu-etal-2012-paraphrasing} to ensure the source data retains distinct stylistic characteristics. Building on this foundation, we design two complex multi-style transformation scenarios: \textbf{(1)} \textit{converting negative Shakespearean text into positive modern English} and \textbf{(2)} \textit{converting positive modern English into negative Shakespearean text}. Table \ref{tab:datasets} presents an overview of five benchmark datasets, including the multi-style transfer dataset.

\begin{figure}
    \centering
    \includegraphics[width=1\linewidth]{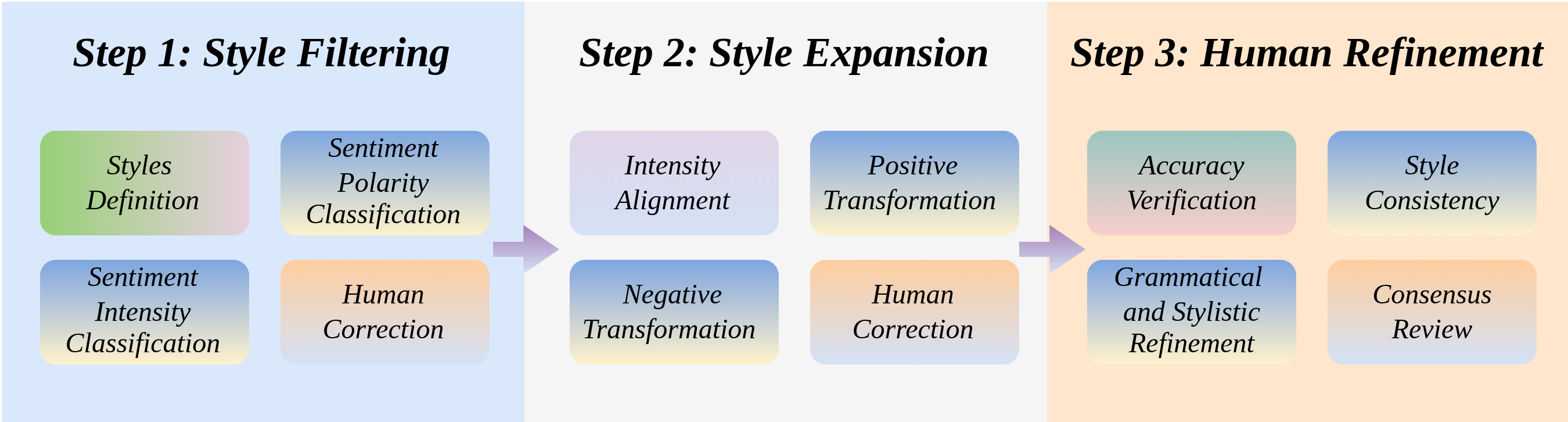}
    \caption{The construction of multi-style benchmark.}
    \label{fig:multi-style}
\end{figure}

Figure \ref{fig:multi-style} illustrates the construction process of our multi-style benchmark. 
Our approach extends beyond simple style transfer and adopts a hierarchical, automated-human collaborative pipeline to ensure the dataset's quality and diversity. Specifically, the construction pipeline consists of the following multi-step style transformation:

\begin{itemize} 
\item \textbf{\textit{Step 1 (Style Filtering)}}: We use roBERTa to perform sentiment polarity classification and emotional intensity classification on sentences from the Shakespearean dataset.
\item \textbf{\textit{Step 2 (Style Expansion)}}: By using the GPT-4~\cite{openai2024gpt4technicalreport} pre-trained model, we apply a negative transformation to the Shakespearean texts and convert the corresponding modern English into positive modern English, ensuring consistent emotional polarity throughout the process.
\item \textbf{\textit{Step 3 (Human Refinement)}}: Human experts review and refine the GPT-4-generated transformations, ensuring accurate stylistic expression, preserving emotional tone, and correcting any discrepancies. 
\end{itemize}

Throughout this process, we construct a high-quality dataset with stylistic combinations.

\subsection{Style Filtering}
In the style filtering step, we define ``negative'' and ``positive'' language styles based on emotional expression. Negative styles convey sentiments such as disappointment, sadness, and despair, often characterized by terms like ``forsaken,'' ``tears,'' and ``grief.'' In contrast, positive styles emphasize hope, encouragement, and happiness, with words such as ``hope,'' ``bright,'' and ``laughter.'' These definitions are informed by established sentiment analysis frameworks, with foundational works like \cite{Pang2008OpinionMA} and \cite{Mihalcea2006ACA} providing the basis for understanding emotional polarity in language, categorizing sentiment into positive, negative, and neutral classes.
To identify these styles, we employed the pre-trained roBERTa model, which excels in grasping both contextual subtleties and emotional undertones. Unlike traditional keyword-based methods, roBERTa captures sentiment with greater precision by accounting for sentence structure and broader context, making it particularly effective for analyzing complex emotional expressions.

We employed roBERTa to analyze the Shakespearean English dataset for sentiment polarity classification and emotional intensity classification. The model automatically classified each sentence as either ``negative'' or ``positive'' and assigned an emotional intensity score on a scale from 1 to 5. These initial automatic classifications were subsequently refined through a thorough manual review process to ensure precision. The final result is a high-quality labeled dataset, comprising sentences with distinct emotional tones, which serves as a robust foundation for future multi-style transfer tasks.

\subsection{Style Expansion}
In the style expansion step, we use the large language model (LLM) for style transfer to achieve two tasks: \textbf{(1)} \textit{converting positive sentences in the Shakespearean English dataset into negative ones}, and \textbf{(2)} \textit{transforming the corresponding modern English sentences into positive modern English}.

First, based on the sentiment intensity labels, we identify the emotional intensity level of each sentence. Using the LLM, we then convert the positive sentences in the Shakespearean English dataset to negative ones, while maintaining a similar emotional intensity. This results in a negative version of the Shakespearean English dataset. Next, we apply the pre-trained GPT-4 model to convert the modern English equivalents of the Shakespearean sentences into positive modern English, ensuring that the emotional intensity of the transformed sentences aligns with the intensity of the original negative Shakespearean sentences.
\subsection{Human Refinement}
While GPT-4 is responsible for generating the initial sentence transformations, human oversight is essential to ensure semantic accuracy and stylistic coherence. To this end, we implemented a manual refinement phase to review and enhance the model's outputs.
\textbf{\textit{Expert annotators}} with backgrounds in literature and linguistics reviewed each transformed sentence to ensure it retained the intended positive tone and original meaning. The process involved the following steps:

\begin{itemize}
    \item \textit{Accuracy Verification}: Annotators confirmed that the transformation accurately preserved the original meaning, correcting any semantic deviations.
    
    \item \textit{Style Consistency}: Annotators ensured the sentences followed a modern, positive style, avoiding outdated or overly contemporary language.
    
    \item \textit{Grammatical and Stylistic Refinement}: Annotators corrected grammatical errors, smoothed awkward phrasing, and selected the most fluent options when multiple alternatives were possible.
    
    \item \textit{Consensus Review}: For complex cases, annotators collaborated to ensure consistency and clarity across the dataset.
\end{itemize}

This rigorous human-in-the-loop process resulted in a high-quality, validated corpus that faithfully reflects both the desired sentiment and stylistic transformation. The combination of expert review and automated generation yielded outputs that are syntactically sound, semantically accurate, and stylistically polished.

\end{document}